%% file: egpaper.tex
\documentclass[10pt,twocolumn,letterpaper]{article}

\usepackage{wacv}
\usepackage{times}
\usepackage{epsfig}
\usepackage{graphicx}
\usepackage{amsmath}
\usepackage{amssymb}

\usepackage[accsupp]{axessibility}  
\usepackage[symbol]{footmisc}

\usepackage[table,xcdraw]{xcolor}
\usepackage{mathtools}
\usepackage{booktabs}
\usepackage{multirow}
\usepackage{arydshln}
\usepackage{makecell}
\usepackage{blindtext}
\usepackage[utf8]{inputenc}
\usepackage{dsfont}
\usepackage{bm}
\usepackage{subcaption,dcolumn}
\usepackage{capt-of}
\usepackage[mode=buildnew]{standalone}
\newcolumntype{d}[1]{D..{#1}}
\makeatletter
\newcommand*{\pmzerodot}{%
  \nfss@text{%
    \sbox0{$\vcenter{}$}
    \sbox2{0}%
    \sbox4{0\/}%
    \ooalign{%
      0\cr
      \hidewidth
      \kern\dimexpr\wd4-\wd2\relax 
      \raise\dimexpr(\ht2-\dp2)/2-\ht0\relax\hbox{%
        \if b\expandafter\@car\f@series\@nil\relax
          \mathversion{bold}%
        \fi
        $\cdot\m@th$%
      }%
      \hidewidth
      \cr
      \vphantom{0}
    }%
  }%
}

\usepackage[pagebackref=true,breaklinks=true,colorlinks,bookmarks=false]{hyperref}

\def\wacvPaperID{742} 

\wacvfinalcopy 

\ifwacvfinal
\fi

\ifwacvfinal
\usepackage[breaklinks=true,bookmarks=false]{hyperref}
\else
\usepackage[pagebackref=true,breaklinks=true,colorlinks,bookmarks=false]{hyperref}
\fi

\begin{document}

\title{A Pixel-Level Meta-Learner\\
       for Weakly Supervised Few-Shot Semantic Segmentation}

\author{Yuan-Hao Lee \qquad Fu-En Yang \qquad Yu-Chiang Frank Wang\\
Graduate Institute of Communication Engineering, National Taiwan University, Taiwan, R.O.C.\\
ASUS Intelligent Cloud Services, Taiwan, R.O.C.\\
{\tt\small \{r07942074, f07942077, ycwang\}@ntu.edu.tw}
}

\maketitle
\thispagestyle{empty}

\input{0_abstract}
\input{1_introduction}
\input{2_related}

\input{3_method}
\input{4_experiments}
\input{5_conclusion}

{\small
\bibliographystyle{ieee_fullname}
\bibliography{egbib}
}

\end{document}

%% file: 0_abstract.tex
\begin{abstract}
Few-shot semantic segmentation addresses the learning task in which only few images with ground truth pixel-level labels are available for the novel classes of interest. One is typically required to collect a large mount of data (i.e., base classes) with such ground truth information, followed by meta-learning strategies to address the above learning task. When only image-level semantic labels can be observed during both training and testing, it is considered as an even more challenging task of weakly supervised few-shot semantic segmentation. To address this problem, we propose a novel meta-learning framework, which predicts pseudo pixel-level segmentation masks from a limited amount of data and their semantic labels. More importantly, our learning scheme further exploits the produced pixel-level information for query image inputs with segmentation guarantees. Thus, our proposed learning model can be viewed as a pixel-level meta-learner. Through extensive experiments on benchmark datasets, we show that our model achieves satisfactory performances under fully supervised settings, yet performs favorably against state-of-the-art methods under weakly supervised settings.
\end{abstract}

%% file: 1_introduction.tex
\section{Introduction}

Recent advances in deep convolutional neural networks (CNNs)~\cite{dcnn} have significantly improved the performances of several computer vision tasks. Among them, \textit{semantic segmentation} aims at predicting class labels for each pixel in an image, with applications ranging from autonomous driving to medical imaging. With the help of CNNs, state-of-the-art semantic segmentation models including FCN~\cite{fcn}, SegNet~\cite{segnet} and DeepLabs~\cite{deeplabv1,deeplabv2,deeplabv3,deeplabv3+} have all achieved very promising results and been successfully applied to the above applications. However, these models are generally trained in a fully supervised manner, requiring a huge amount of training data with pixel-level annotation for each category of interest. This substantially limits the scalability and practicality of these models, as collecting densely labeled data would be very time-consuming.

\begin{figure}
  \centering
  \includegraphics[width=0.91\linewidth, angle=0]{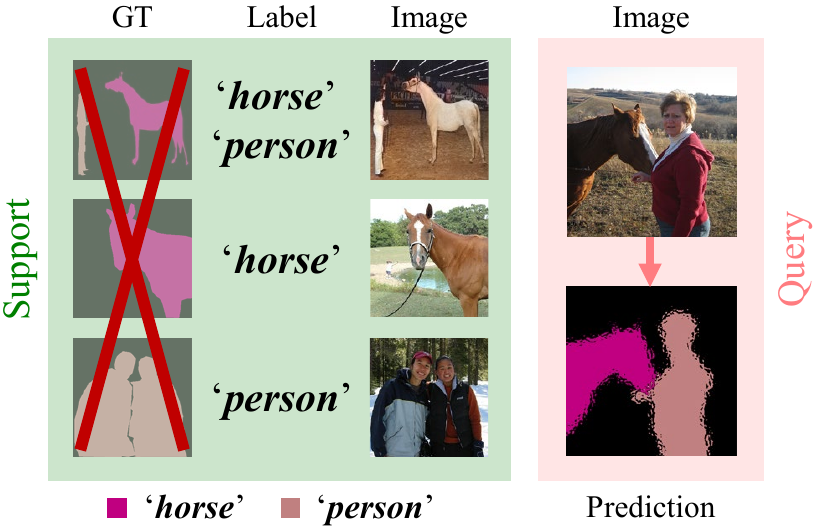}
  \caption{Illustration of weakly supervised few-shot segmentation. With only semantic labels but not pixel-level masks observed during few-shot training and testing, semantic segmentation of particular image categories can be achieved. Note that a 2-way 3-shot scheme is depicted.}
  \label{fig:intro}
\end{figure}

Extended from the learning task of semantic segmentation and few-shot learning, \emph{few-shot semantic segmentation} considers a more challenging setting in which only a few images are with ground-truth pixel-level labels for the (novel) classes of interest. In order to realize the learning of few-shot segmentation models, existing methods typically adopt meta-learning schemes~\cite{meta1,meta2,meta3}, utilizing support and query images sampled from base categories (i.e., those with a sufficient amount of training data) for performing pixel-wise classification. Recent methods like AMP~\cite{amp}, PANet~\cite{panet}, FWB~\cite{FWB}, PFENet~\cite{pfenet} and ASGNet~\cite{asgnet} choose to extract prototypes from support-set images using their ground truth masks, and expect such prototypes to sufficiently describe the associated semantic category. Nevertheless, these existing methods require pixel-level ground truth labels for each image of the base categories, and cannot be easily extended to weakly supervised settings.

In order to alleviate the requirement of annotating pixel-level ground truth label information, we consider an even more challenging yet practical setting of weakly supervised few-shot segmentation, which requires only image-level labels collected for images in both base and novel categories, as depicted in Figure~\ref{fig:intro}. As shown in the figure, we aim at inferring pixel-level labels from image-level labels in few-shot settings, followed by a meta-learning scheme enforced at pixel-level for segmentation purposes. 

With this goal in mind, we propose a novel learning scheme in this paper, focusing on deriving a pixel-level meta learner for weakly supervised few-shot semantic segmentation. During the meta-training stage, our proposed learning framework observes only image-level labels and utilizes Classification Activation Maps (CAM)~\cite{cam,gradcam} for prediction of pseudo pixel-level labels for each input image. While the class label embedding~\cite{nnlm,word2vec} is exploited to bridge the gap between image and pixel-level information, our proposed method does not encounter any information leak since the class labels of both base and novel categories are \textit{not} present in those of CAM.

Under the guidance of the produced pseudo pixel-level labels, we uniquely reinterpret the original problem as a \emph{pixel-wise few-shot classification} task. That is, we view each pixel in support/query set images as individual samples, and turn the proposed model into a pixel-level meta learner for few-shot semantic segmentation. As confirmed later by our experiments, our model not only achieves satisfactory performances on standard fully supervised few-shot semantic segmentation tasks, it would perform favorably against several state-of-the-art approaches on benchmark datasets under weakly supervised settings.

The contributions of this work are summarized below:
\begin{itemize}
\item We address weakly supervised few-shot semantic segmentation, which requires only image-level labels during both training and testing.
\item We bridge the gap between image and pixel-level labels using classification activation maps with no information leak, which allows prediction of pseudo pixel-level labels in weakly supervised settings.
\item We uniquely design a pixel-level meta learner which enforces segmentation consistency across support and query set images during few-shot semantic segmentation. The proposed model can be realized in both fully supervised and weakly supervised settings.
\end{itemize}

\begin{figure*}
  \centering
  \includegraphics[width=\textwidth, angle=0]{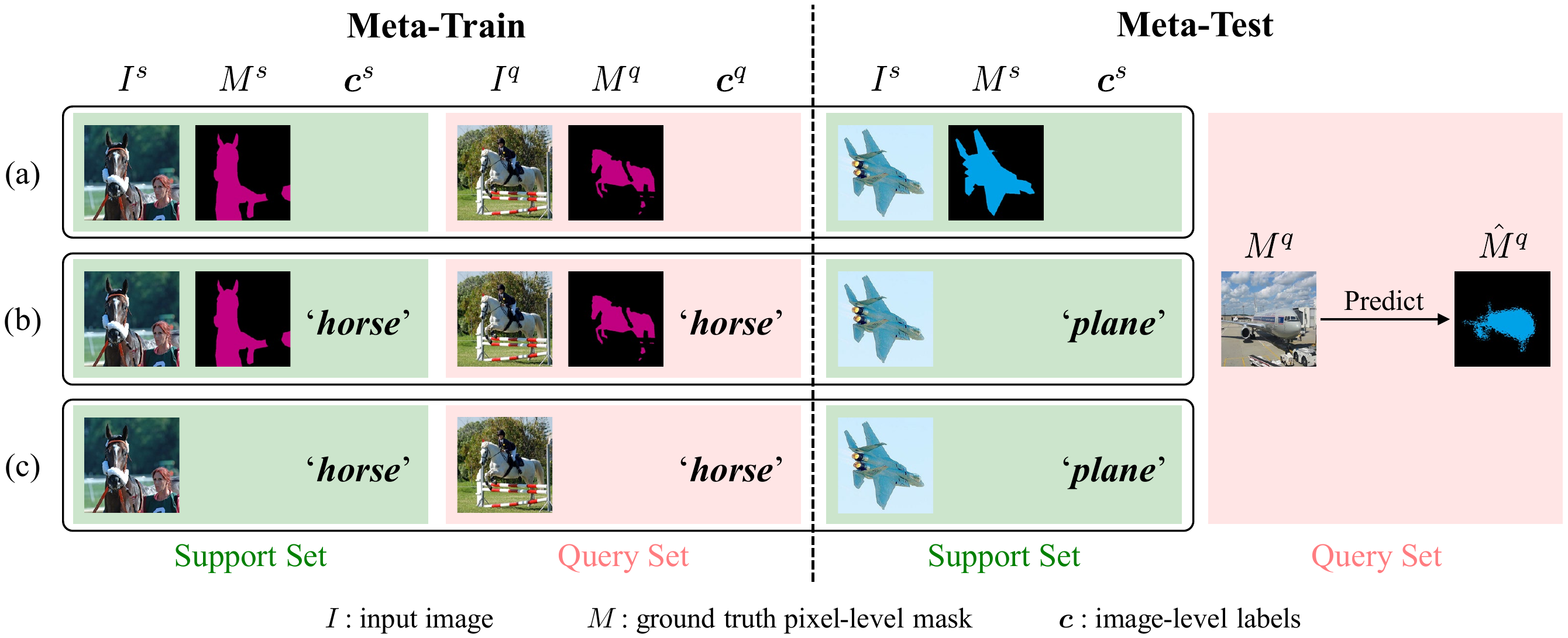}
  \caption{Comparisons of different \emph{few-shot} semantic segmentation schemes. (a) \emph{Fully Supervised}~\cite{amp,panet,canet,FWB,pfenet}: $M$ required in meta-training and meta-testing; (b) \emph{Loosely Weakly Supervised}~\cite{coattention}: both $M$ and $c$ available during meta-training, while only $c$ observed in meta-testing; (c) \emph{Weakly Supervised}: only image-level labels $c$ available in both phases.}
  \label{fig:fig1}
\end{figure*}

%% file: 2_related.tex
\section{Related Works}

\noindent\textbf{Semantic Segmentation.}
The task of semantic segmentation aims at performing pixel-level classification for each image. Most recent methods involve the use of deep convolutional neural networks (e.g., FCN~\cite{fcn}). Following works include SegNet~\cite{segnet}, PSPNet~\cite{pspnet}, U-Net~\cite{unet}, DeepLabs~\cite{deeplabv1,deeplabv2,deeplabv3,deeplabv3+} and FastFCN~\cite{fastrcn}. A notable improvement is achieved by embedding features that contain multi-scale context information by applying dilated convolution~\cite{multiscale,deeplabv1} or spatial pyramid pooling~\cite{pspnet,deeplabv2}. These methods, however, are trained in a fully supervised manner and require a huge amount of pixel-level labels, which are time-consuming and laborious to obtain.

\noindent\textbf{Weakly Supervised Semantic Segmentation.} To alleviate the need for densely annotated ground truth data, some recent works address semantic segmentation in weak supervision, which utilize multiple-instance learning~\cite{mil1,mil2}, graph~\cite{graph1,graph2,graph3} and self-training based~\cite{self1,self2,self3,edam} techniques. However, such weakly supervised methods cannot be easily extended to few-data or open-set scenarios.


\noindent\textbf{Few-Shot Semantic Segmentation.}
Few-shot learning aims at learning models which would generalize to categories with only a limited amount of labeled data~\cite{fewshot1,fewshot2}. \textit{Meta-learning}~\cite{meta1,meta2,meta3} has been widely applied for this task, with the core idea of adapting the learning scheme from base to novel categories. For example, metric-based meta-learning algorithms such as ProtoNet~\cite{protonet} and RN~\cite{rn} learn feature embeddings that exhibit proper distance metrics for classification and generalization. \textit{Few-shot semantic segmentation}, on the other hand, aims at generalizing the ability of pixel-level classification across categories, while only a limited number of images are with ground truth pixel-level labels. OSLSM~\cite{oslsm} is the first proposed method to tackle this problem, leveraging information learned from support-set images and outputs parameters for query image segmentation. PL~\cite{pl} adopts metric learning methods~\cite{protonet} to extract prototypes of each semantic class, and measures their distances between feature maps of query images; CANet~\cite{canet} adds an iterative optimization module to refine the predicted results; PFENet~\cite{pfenet} generates additional prior masks to enrich the extracted features. \newline
\indent Moreover, PANet~\cite{panet}, FWB~\cite{FWB} and CRNet~\cite{crnet} propose to further leverage information from the support set by performing segmentation in the reversed direction (i.e., segmentation of the support set) for improved model learning. To exploit knowledge from the foreground objects, DAN~\cite{dan} and SimPropNet~\cite{simpropnet} introduce attention mechanisms, PMMs~\cite{pmms} and ASGNet~\cite{asgnet} employ multiple prototypes for a single category, and PPNet~\cite{ppnet} proposes part-aware prototypes to capture fine-grained features. While some of the existing few-shot semantic segmentation methods present results in weak supervision settings (e.g., use of bounding boxes or scribbles~\cite{cofcn1,panet,canet} as guidance), they cannot produce satisfactory performance with only image-level annotation observed. More recently, \cite{weaklyoneshot,zero,coattention} follow the few-shot setting using image-level supervision, but they still require collection of ground truth pixel-level masks for base-class images during meta-training. As depicted in Figure~\ref{fig:fig1} and Table~\ref{tab:comparison}, to the best of our knowledge, we are the first to tackle few-shot semantic segmentation using only image-level annotations during both (meta) training and testing stages.


\begin{figure*}
  \centering
  \includegraphics[width=\textwidth, angle=0]{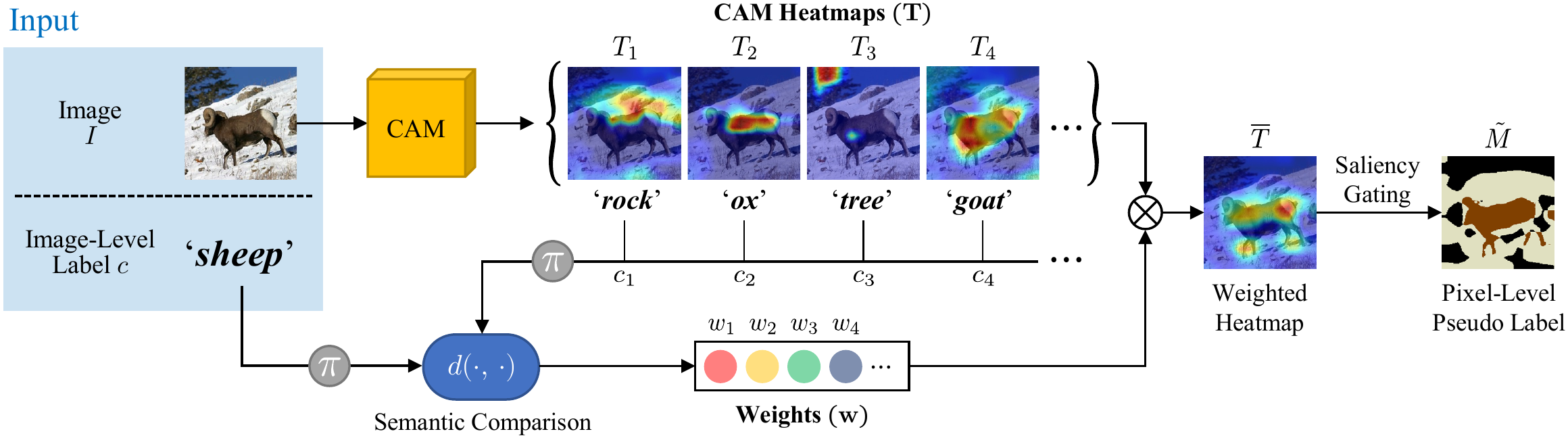}
  \caption{Illustration of pixel-level pseudo label generation. Given an input $I$ and its image-level label $c$, the CAM module extracts heatmaps $\mathbf{T}$ for each training class $c_k$ of CAM (not in $\mathcal{C}_\text{base}$ nor $\mathcal{C}_\text{novel}$). $\mathbf{w}$ denotes the visual similarity for the heatmap of each CAM category, which is measured by the distance between the word embeddings of the associated class labels. The output heatmap $\overline{T}$ is converted into the pseudo mask $\tilde{M}$ for $I$ via saliency gating. Note that $\pi$ represents the word embedding function. }
  \label{fig:g}
\end{figure*}

\input{tabs/comparison}

%% file: tabs/comparison.tex
\begin{table}[]
  \centering
  \resizebox{.9\linewidth}{!}{%
  \begin{tabular}{ccc} 
    \hline\hline
    Method & {Weak Supervision} & {Few-Shot Setting} \\
    \Xhline{2\arrayrulewidth}
    DeepLab~\cite{deeplabv1}  & - & - \\
    EDAM~\cite{edam}          & $\checkmark$ & - \\
    PFENet~\cite{pfenet}      & - & $\checkmark$ \\
    Co-att~\cite{coattention} & $\triangle$ & $\checkmark$ \\
    \hline
    \textbf{Ours}             & $\checkmark$ & $\checkmark$ \\
    \hline\hline
    \\
  \end{tabular}
  }
  \caption{Comparisons of different semantic segmentation methods. Existing methods cope with either weak supervision or few-shot settings, while ours combine both. Note that \cite{coattention} still requires full supervision during training, as detailed in Figure~\ref{fig:fig1}(b).}
  \label{tab:comparison}
\end{table}

%% file: 3_method.tex
\section{Proposed Method}

\subsection{Notation and Problem Formulation}\label{sec:notation}
For the sake of completeness, we define the notations which will be used in this paper. The semantic classes are denoted as $\mathcal{C}$, which are split into two disjoint subsets: base categories $\mathcal{C}_\text{base}$ and novel categories $\mathcal{C}_\text{novel}$ (i.e., $\mathcal{C}_\text{base} \cup \mathcal{C}_\text{novel} = \mathcal{C}$ and $\mathcal{C}_\text{base} \cap \mathcal{C}_\text{novel} = \varnothing$). Note that the novel categories are with only a few samples available during training (typically less than 5 per class). For each input RGB image $I\in \mathbb{R}^{H\times W \times 3}$, we denote its image-level labels as $\bm{c}\subseteq \mathcal{C}$, and the associated ground truth pixel-level semantic mask as $M\in \{\bm{c}\cup \pmzerodot\}^{H\times W}$, where $\pmzerodot$ indicates the background pixels. For each training episode, we sample a support/query pair from the image dataset $\mathcal{D}_\text{train} = \{(S_j, Q_j)\}^{n_\text{train}}_{j=1}$ that contain the same set of base categories, while the testing episodes consist of those from $\mathcal{C}_\text{novel}$ (i.e., $\mathcal{D}_\text{test} = \{(S_j, Q_j)\}^{n_\text{test}}_{j=1}$). 

\begin{figure*}
  \centering
  \includegraphics[width=\textwidth, angle=0]{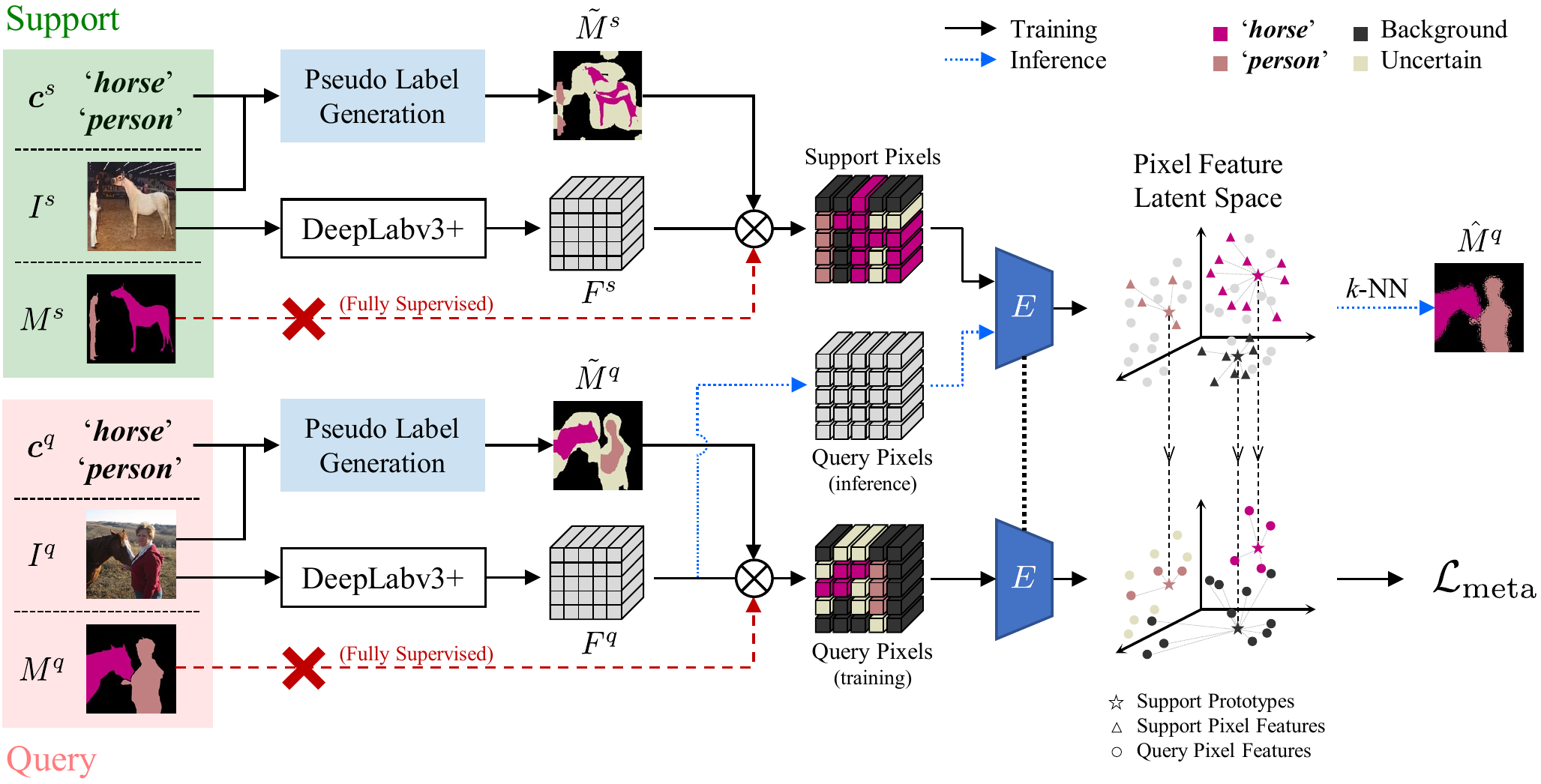}
  \vspace{-0.2cm}
  \caption{Architecture of our pixel-level meta-learner for weakly supervised segmentation. During (meta) training, DeepLabv3+ extracts pixel-wise feature maps $F^s$ and $F^q$ for support and query inputs, with the produced pseudo pixel-level labels $\tilde{M}^s$ and $\tilde{M}^q$, respectively. Our meta-learner encoder $E$ transforms the above $F$ into a latent space, in which the pixel-level prototypical loss $\mathcal{L}_\text{meta}$ can be calculated based on $\tilde{M}$ for segmentation purposes. During inference (i.e., meta-testing), segmentation $\hat{M}^q$ for the query can be performed by pixel-wise $k$-NN classification using $F^s$ and $\tilde{M}^s$. 
  }
  \label{fig:main}
\end{figure*}

In standard fully supervised $N$-way $K$-shot settings, each support set $S_j = \{(I^s_i,M^s_i)\}^{N\times K}_{i=1}$ contains $N\times K$ image/label example pairs ($K$ pairs from each of $N$ categories), and the query sets are denoted as $Q_j = \{(I^q_i,M^q_i)\}^{n_q}_{i=1}$ where $n_q$ is the number of query images in a single episode. As for the \textit{weakly supervised} setting considered in this work, only image-level labels are available during training and testing, so that support/query sets are denoted as $S_j = \{(I^s_i,\bm{c}^s_i)\}^{N\times K}_{i=1}$ and $Q_j = \{(I^q_i,\bm{c}^q_i)\}^{n_q}_{i=1}$, respectively. Our proposed weakly supervised learning framework is able to produce pseudo masks of each image given only its class label (as depicted in Figure~\ref{fig:g}), followed by our pixel-level meta-learner for few-shot semantic segmentation (see Figure~\ref{fig:main} for the complete framework).


\subsection{Pixel-Level Pseudo Label Generation}\label{sec:pseudogen}

In the weakly supervised scenario, only the class labels are available for the image data during both training and inference. Thus, we first present a module that is able to generate pseudo pixel-level semantic masks, guiding the following segmentation process. As shown in Figure~\ref{fig:g}, this pseudo pixel-level label generation can be viewed as the process of semantics-oriented heatmap extraction. As adopted in previous works like~\cite{huang2018weakly}, Classification Activation Maps (CAM)~\cite{cam} have been utilized to localize discriminative regions in images that are informative in classifying image-level labels. In order to produce the heatmap for the input image $I$ as its pseudo pixel-level labels, we follow~\cite{panet,canet} and apply a VGG-16~\cite{vgg16} network as our CAM backbone. It is worth noting that, the CAM backbone is pre-trained on a reduced subset of ImageNet~\cite{imagenet}, in which the images do not belong to either the base or novel categories in our segmentation task. This would minimize possible leakage of semantic information.

With CAM obtained, we extract per-class heatmaps $\mathbf{T}$ of the input image $I$:

\begin{equation}\label{eq:cam}
  \begin{aligned}
  \text{CAM}(I)=\mathbf{T}=[T_1,T_2,\ldots,T_{N_\text{CAM}}]\text{,} \\
  \end{aligned}
\end{equation}

\noindent where $T_k\in [0,1]^{H\times W}$ denotes the heatmap of the $k$-th image class in CAM, whereas $N_\text{CAM}$ is the total number of pre-training image categories. 

With the above per-class heatmaps observed, we next leverage the word embedding features of each class label as intermediate representations, with the associated similarity indicating the weight for each of the $N_\text{CAM}$ categories. More specifically, as depicted in Figure~\ref{fig:g}, we extract word embedding features of each class label in CAM and that of the input image. We perform pairwise comparisons between these features to obtain weighting factors $w_k$ for each CAM category $c_k$:

\begin{equation}\label{eq:wk}
  \begin{aligned}
  w_k = d(\pi(c), \pi(c_k))^{-1}\text{,} \\
  \end{aligned}
\end{equation}

\noindent where $\pi(\cdot)$ represents the word embedding function. We note that $d(\cdot,\cdot)$ denotes the distance metric, and we use consine similarity in our work. Thus, categories that are semantically similar with each other (e.g., \emph{goat}/\emph{sheep}) would result in a higher weight, and vice versa for those that are dissimilar (e.g., \emph{goat}/\emph{tree}). As a result, a weighted heatmap $\overline{T}$ can be obtained by averaging the above $N_\text{CAM}$ heatmaps, which is calculated as

\begin{equation}\label{eq:w}
  \begin{aligned}
  \overline{T} = \mathbf{w}\cdot \mathbf{T}^\top = \sum\limits_{k=1}^{N_\text{CAM}} w_k T_k\text{.} \\
  \end{aligned}
\end{equation}

As the CAM heatmaps identify only regions that are helpful in terms of classification, more detailed structural information such as edges and boundaries would not be well described. To this end, we impose a class-agnostic saliency map on the weighted heatmap as a gating mechanism, alleviating the presence of false-positive pixels in the generated pseudo labels. Here we note that, to comply with our weakly supervised setting, the saliency maps are obtained via a network pre-trained with only foreground/background information without any categorical supervision. In other words, minimized leakage of semantic information is also enforced at this stage.





\subsection{Pixel-Level Meta-Learner for Few-Shot Semantic Segmentation}\label{sec:main}

We now detail our proposed meta-learning scheme for few-shot semantic segmentation. Under the guidance of CAM, the pseudo pixel-level labels obtained in Section~\ref{sec:pseudogen} tend to contain only partial discriminative areas, and may not sufficiently cover the entire foreground object. In order to learn few-shot segmentation models with only such information from pseudo labels, we present a unique pixel-level meta-learning framework, as illustrated in Figure~\ref{fig:main}. In our pixel-level meta-learner, we utilize DeepLabv3+~\cite{deeplabv3+} as the feature extractor, together with the introduced pseudo label generation and learning modules for pixel-wise classification. It is worth noting that, the DeepLabv3+ is pre-trained on categories \textit{not} appeared in $\mathcal{C}$, and this module remains fixed throughout the meta-learning process. In other words, no semantic information is leaked from its training stage. By removing the final classification layers of DeepLabv3+, we obtain a pixel-wise feature map $F^s\in \mathbb{R}^{H\times W\times d}$ from the support image $I^s$, where $(H, W)$ is the original image dimension and $d$ is the feature channel size.


Next, we label each spatial location in $F^s$ using the generated support pseudo mask $\tilde{M}^s$. By randomly sampling a fixed number of pixels from each background/foreground category, the resulting pixel-wise features are then collected into a set of \emph{support pixel features}. It is worth noting that, we choose not to use all the labeled pixel features. This is not only because that such pseudo labels might not match the ground truth ones (although not available), this sampling mechanism also makes the meta-learning process more robust against weak labels.

Likewise, a pixel-wise feature map $F^q\in \mathbb{R}^{H\times W\times d}$ of the query image $I^q$ is also obtained via the same DeepLabv3+ feature extractor, resulting in a total of $H\times W$ $d$-dimensional \emph{query pixel features}. All support and query pixel features are then jointly embedded into a latent space via the learnable encoder $E$.
With the support pixel features and pseudo labels obtained, we define the prototypes $p_c$ for each associated category $c\in\{\bm{c}^s\cup \pmzerodot\}$ as

\begin{equation}\label{eq:proto}
  \begin{aligned}
  p_c = \frac{\sum_{l}E(F^s_{l})\mathds{1}[\tilde{M}^s_{l}=c]}{\sum_{l}\mathds{1}[\tilde{M}^s_{l}=c]}\text{,} \\
  \end{aligned}
\end{equation}

\input{tabs/pascal_1shot}

\noindent where $l$ iterates over each pixel, and $\mathds{1}[\cdot]$ is an indicator function which only outputs $1$ when the condition holds. 

Inspired by~\cite{protonet, pl, pfenet}, we advance the prototypical loss $\mathcal{L}_\text{meta}$ as the objective during the meta-training stage, which is calculated by accumulating the distance between each query pixel (with pseudo labels) and its corresponding pixel-level prototype from the support sets:

\begin{equation}\label{eq:Lseg}
  \begin{aligned}
  \mathcal{L}_\text{meta} = -\frac{\sum_{c}\sum_{l}\exp(-d(E(F^q_{l}), p_{c}))\mathds{1}[\tilde{M}^q_{l}=c]}{\sum_{c}\sum_{l}\mathds{1}[\tilde{M}^q_{l}=c]}\text{.} \\  
  \end{aligned}
\end{equation}

During inference (i.e., meta-testing), the labels of the embedded query pixel features are determined by performing pixel-wise $k$-nearest neighbors classification, as depicted by the blue dotted arrows in Figure~\ref{fig:main}. The final predicted query semantic mask $\hat{M}^q$ is then compared with the ground truth $M^q$ for performance evaluation.


%% file: tabs/pascal_1shot.tex
\begin{table*}[!tp]
    \centering
    \resizebox{.92\linewidth}{!}{%
    \begin{tabular}{cccccccccc} 
        \hline\hline
        \multicolumn{2}{c}{(\textit{fully sup.} / \textit{weakly sup.})} & \multicolumn{6}{c}{\textbf{1-shot}} & \multicolumn{2}{c}{\textbf{5-shot}} \\ 
        \cmidrule(lr){3-8}\cmidrule(lr){9-10}
        Method & Backbone & Split-0  & Split-1  & Split-2  & Split-3  & Mean & $\Delta$ & Mean & $\Delta$               \\ 
        \Xhline{2\arrayrulewidth}
        \textbf{1-way} \\
        Co-att~\cite{coattention} & VGG-16 & 49.5 & 65.5 & 50.0 & 49.2 & 53.5 & --- & 51.7 & --- \\
        AMP~\cite{amp} & VGG-16 & 41.9 / 10.6 & 50.2 / 14.1 & 46.7 / 7.6 & 34.7 / 10.9 & 43.4 / 10.8 & 32.6 & 46.9 / 14.7 & 32.2 \\
        PANet~\cite{panet} & VGG-16 & 42.3 / 25.7 & 58.0 / 33.4 & 51.1 / 28.8 & 41.2 / 20.7 & 48.1 / 27.1 & 21.0 & 55.7 / 37.7 & 18.0 \\
        PFENet~\cite{pfenet} & ResNet-50 & 61.7 / 33.4 & 69.5 / 42.5 & 55.4 / 43.6 & 56.3 / 39.9 & 60.8 / 39.9 & 20.9 & 61.9 / 44.8 & 17.1 \\
        Ours & VGG-16 & 38.3 / 36.5 & 57.6 / 51.7 & 54.0 / 45.9 & 40.1 / 35.6 & 47.5 / \textbf{42.4} & \textbf{5.1} & 50.6 / \textbf{45.5} & \textbf{5.1} \\
        \hline
        \textbf{2-way} \\
        PANet~\cite{panet} & VGG-16 & 45.1$^\dagger$ / 24.5 & 45.1$^\dagger$ / 33.6 & 45.1$^\dagger$ / 26.3 & 45.1$^\dagger$ / 20.3 & 45.1 / 26.2 & 18.9 & 53.1 / 36.6 & 16.5 \\
        Ours & VGG-16 & 36.5 / 31.5 & 51.8 / 46.7 & 48.5 / 41.4 & 38.9 / 31.2 & 43.9 / \textbf{37.7} & \textbf{6.2} & 49.3 / \textbf{43.0} & \textbf{6.3} \\
        \hline\hline
    \end{tabular}
    }

    (a) PASCAL-$5^i$

    \resizebox{.885\linewidth}{!}{%
    \begin{tabular}{cccccccccc}
        {} & {} & {} & {} & {} & {} & {} & {} & {} & {} \\ 
        \hline\hline
        \multicolumn{2}{c}{(\textit{fully sup.} / \textit{weakly sup.})} & \multicolumn{6}{c}{\textbf{1-shot}} & \multicolumn{2}{c}{\textbf{5-shot}} \\ 
        \cmidrule(lr){3-8}\cmidrule(lr){9-10}
        Method & Backbone & Split-0  & Split-1  & Split-2  & Split-3  & Mean & $\Delta$ & Mean & $\Delta$               \\ 
        \Xhline{2\arrayrulewidth}
        \textbf{1-way} & \phantom{ResNet-50} \\
        PANet~\cite{panet} & VGG-16 & 20.9$^\dagger$ / 12.7 & 20.9$^\dagger$ / 8.7 & 20.9$^\dagger$ / 5.9 & 20.9$^\dagger$ / 4.8 & 20.9 / 8.0 & 12.9 & 29.7 / 13.9 & 15.8 \\
        Ours & VGG-16 & 26.0 / 24.2 & 14.5 / 12.9 & 20.0 / 17.0 & 18.3 / 14.0 & 19.7 / \textbf{17.0} & \textbf{2.7} & 27.0 / \textbf{17.5} & \textbf{9.5} \\
        \hline
        \textbf{2-way} \\
        Ours & VGG-16 & 18.2 / 17.4 & 12.2 / 9.5 & 9.1 / 10.4 & 6.5 / 7.1 & 11.5 / \textbf{11.1} & \textbf{0.4} & 14.8 / \textbf{11.9} & \textbf{2.9} \\
        \hline\hline
    \end{tabular}
    }

    (b) MS COCO\phantom{$^i$}

    \caption{Performance evaluation on (a) PASCAL-$5^i$ and (b) MS COCO in terms of mean-IoU (Mean) and performance difference $\Delta$ due to change of settings. The numbers before and after `/' indicate results under fully and weakly supervised settings, respectively. Note that \cite{coattention} considers a loosely weakly supervised setting and requires ground truth pixel-level masks during training, while \cite{pfenet} utilizes a stronger backbone (ResNet-50) compared to others (VGG-16).
    }
    \label{tab:pascal_coco}
\end{table*}

%% file: 4_experiments.tex
\section{Experiments}

\noindent\textbf{Datasets.} We follow the evaluation protocol in~\cite{oslsm} and conduct experiments on the PASCAL-$5^i$ dataset. It contains a total of 20 object categories from the PASCAL VOC 2012~\cite{pascalvoc} and the extended SDS~\cite{sds} datasets, which are evenly divided into 4 splits ($i=0,1,2,3$). Additionally, we consider the MS COCO 2014~\cite{mscoco} dataset, which contains 80 object categories and thus is more challenging. Following the settings of~\cite{panet}, we divide MS COCO into 4 splits with 20 categories each. For both datasets, three of the splits are used as base (training) classes, with the remaining one as novel (testing) classes in each experiment.

\noindent\textbf{Evaluation Metrics.} 
We follow~\cite{oslsm,sgone,panet,canet,FWB} and apply the mean intersection over union (mean-IoU) as the evaluation metric, which computes separate IoUs for each foreground category, and then averages them along with the background class. Following the protocol in~\cite{panet}, the mean-IoU in each evaluation is calculated by the average of 5 runs, with each run randomly sampling 1,000 episodes from the testing set.

\begin{figure*}
  \centering
  \includegraphics[width=.77\textwidth, angle=0]{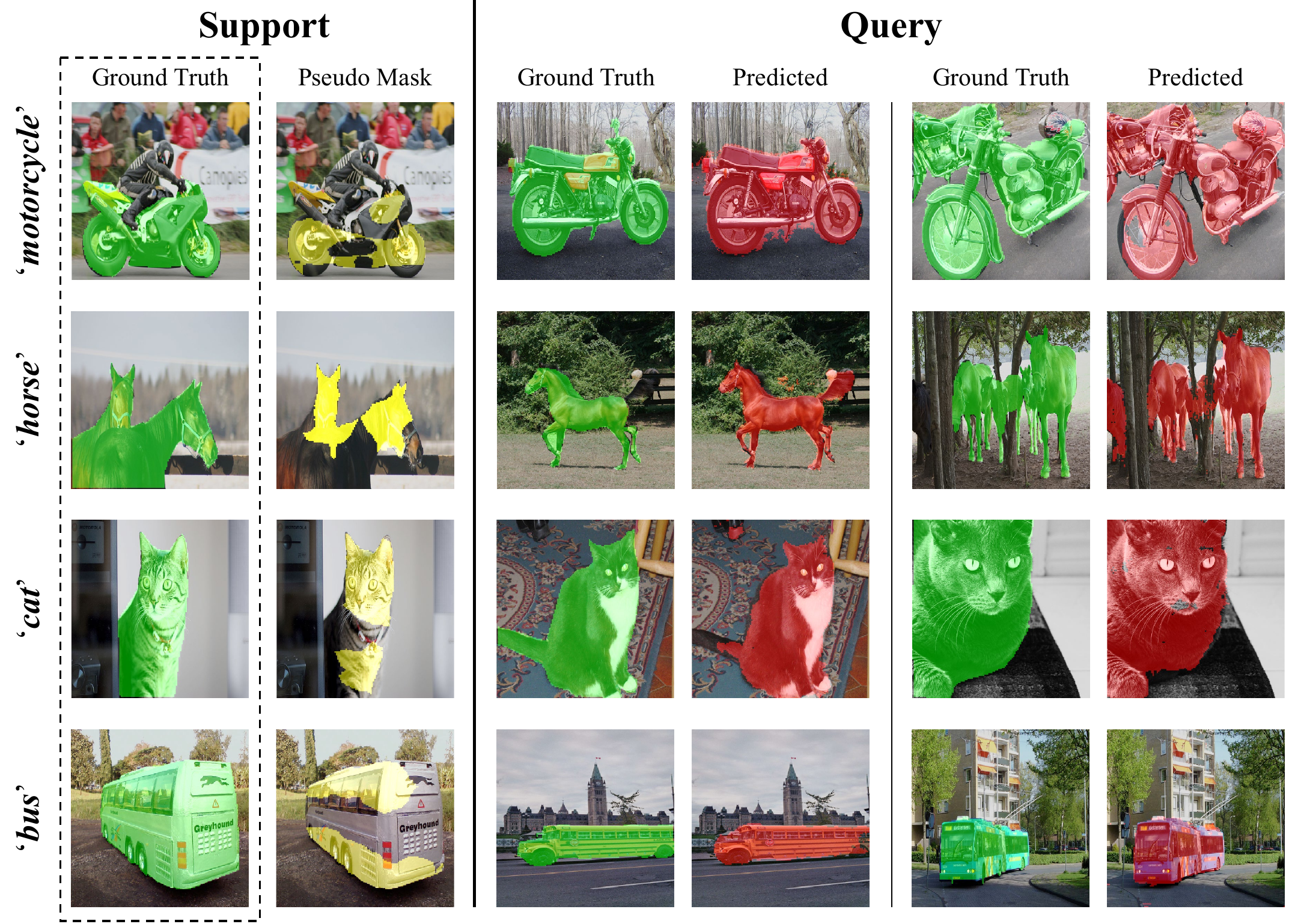}
  \caption{Example 1-way 1-shot segmentation results on PASCAL-$5^i$. We have green, yellow and red pixels represent ground truth pixel-level foreground labels, generated pseudo masks, and the predicted outputs, respectively. Note that our weakly supervised setting does not observe ground truth pixel-level labels for the support set images (in dotted frames).}
  \label{fig:vis}
\end{figure*}
\footnotetext[2]{Split-wise results not reported in the original paper.}

\noindent\textbf{Implementation Details.} In our experiments, all images are normalized and reshaped to $129\times 129$ pixels. For pseudo pixel-level label generation, we choose the VGG-16~\cite{vgg16} network as the CAM backbone, with weights pre-trained over a reduced subset of ILSVRC 2012~\cite{imagenet} with categories in $\mathcal{C}_\text{base}$ and $\mathcal{C}_\text{novel}$ removed. For semantic labels, we use the word embedding vectors pre-trained on Wikipedia using fastText~\cite{fasttext}.
We apply DSS~\cite{saliency2} to extract saliency maps for the gating mechanism in Fig.~\ref{fig:g}, which is a model pre-trained over MSRA-B~\cite{msrab} without categorical supervision.

For the segmentation model, we use a DeepLabv3+~\cite{deeplabv3+} pre-trained over irrelevant categories, which remains fixed throughout the meta-learning process.
The encoder $E$ is a multilayer perceptron with two hidden layers, and the output dimension (for the latent space) is set to 64.
All experiments are implemented by PyTorch, and are run using a single NVIDIA Titan RTX graphics card with 24GB of video memory.

\subsection{Comparison with State-of-the-art Methods}\label{sec:comp}

Since no previous work was designed to address this weakly supervised few-shot segmentation task, we choose to implement modified versions of state-of-the-art supervised methods by replacing the ground truth masks for training with our generated pseudo labels (i.e., as introduced in Section~\ref{sec:pseudogen}). As depicted in Figure~\ref{fig:main}, our proposed framework can be trained in a fully supervised fashion (i.e., using ground truth pixel-level masks during training). Thus, we include this fully supervised version of our model as the performance upper bounds. We compare our results mainly with methods using the same backbone~\cite{coattention,amp,panet}, with the exception of \cite{pfenet} for cross-backbone analysis.

\noindent\textbf{PASCAL-$\bm{5^i}$.} We first compare the performances of different methods on PASCAL-$5^i$, with results listed in Table~\ref{tab:pascal_coco}a. In the first row of this table, we consider a loosely weakly supervised model of \cite{coattention} (as illustrated in Figure~\ref{fig:fig1}b). For other methods (including ours) in this table, we present results under both fully and weakly supervised settings. While our fully supervised model achieved comparable performance with state-of-the-art methods in the standard 1-way 1-shot setting, our model reported a significant improvement over PANet \cite{panet} by 15.3\% (42.4\% v.s. 27.1\%) in the weakly supervised setting. It is worth noting that PFENet~\cite{pfenet} is trained using a stronger backbone (ResNet-50), while all other methods (including ours) utilize VGG-16. Nevertheless, our model still outperforms \cite{pfenet} by a considerable margin of 2.5\% in the weakly supervised setting. We also observe that the performance drop between the two different settings of our model is significantly less than the others. That is, when only image-level labels (instead of ground truth pixel-level masks) are observed during both training and testing, both PFENet~\cite{pfenet} and PANet~\cite{panet} suffered from a $>$20\% performance drop while only 5.1\% was reported by our model.

\noindent\textbf{MS COCO.} As shown in Table~\ref{tab:pascal_coco}b, despite the increased difficulty in few-shot segmentation on MS COCO, our model is able to achieve satisfactory performances under both fully and weakly supervised settings when comparing to \cite{panet}. Specifically, we only observe a 2.7\% performance drop between the two settings on the 1-way 1-shot task, while that of \cite{panet} is 12.9\%. The above quantitative results support the use of our propose framework for solving few-shot semantic segmentation, especially when only image-level labels can be observed during both training and testing (i.e., the weakly supervised setting).
 
\noindent\textbf{Multi-way Segmentation.} We now show that our model is applicable to the cases when there is more than one foreground object category in an image, which is more challenging since it requires more information to be learned in each episode. In the lower parts of Table~\ref{tab:pascal_coco}, we list the performances of different methods under the 2-way setting (i.e., two types of foreground objects exist in an image). It is worth noting that, while most existing methods~\cite{oslsm,amp,canet,FWB,hu2019attention,pfenet} are designed to tackle only 1-way segmentation, they typically claim such extension can be realized by forward passing $K$ times with additional decision rules. On the contrary, our method can directly produce output labels of a multi-category image as the final classification by a $k$-NN search. As shown in Table~\ref{tab:pascal_coco}a, our model performed favorably against previous methods by a margin of 11.5\% on the PASCAL-$5^i$ 2-way 1-shot task. To the best of our knowledge, we are the first to report results of 2-way tasks for the MS COCO dataset, as shown in the last row of Table~\ref{tab:pascal_coco}b.


\begin{figure}
  \centering
  \includegraphics[width=.83\linewidth, angle=0]{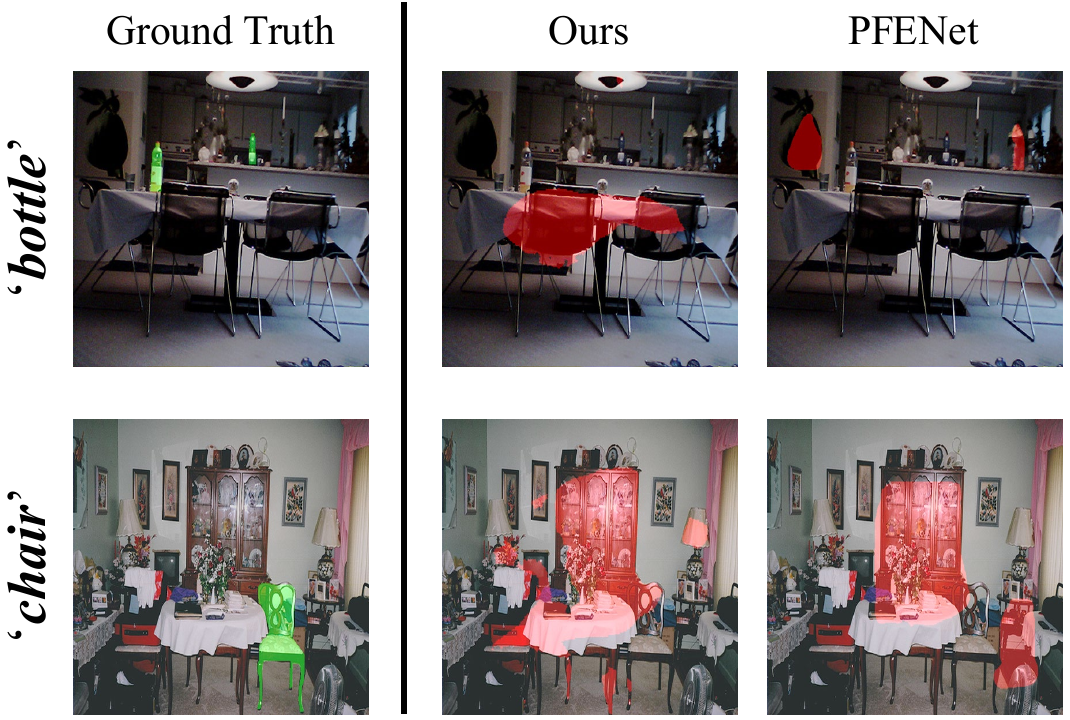}
  \caption{Example failure cases for our weakly supervised model and the fully supervised PFENet~\cite{pfenet}.}
  \label{fig:fail}
\end{figure}

\subsection{Analysis of Our Proposed Method}\label{sec:ablation}


\noindent\textbf{Ablation Study.} As our meta-learning process is mainly achieved by the episodic learning of encoder $E$, we now design a baseline version such that it directly predicts the output mask by a $k$-NN search on the pixel features encoded by the DeepLabv3+ backbone (i.e., without embedding into the latent space via $E$). As shown in the first row of Table~\ref{tab:ablation}, the results were severely degraded with a drop of up to 18\%, if the model was not learned via the meta-learning objectives. This further confirms that, the promising performance achieved by our proposed method is not a direct result of using particular strong backbones. Instead, it leverages spatial and structural details captured by the backbone, upon which semantic and category-wise information is reinforced via the meta-learning process. 

Additionally, we provide results that are trained using pseudo masks generated without saliency gating (as mentioned in the last step of Section~\ref{sec:pseudogen}). As shown in the second row of Table~\ref{tab:ablation}, a slight decrease in mean-IoU (less than 5\%) was observed, while still outperforming the baseline version by a large margin. Thus, the use of saliency gating as the post-processing step for pseudo pixel-level masks would be preferable but not critical.

\input{tabs/ablation}

\noindent\textbf{Qualitative Analysis.} For visual comparisons, we present qualitative results of 1-way 1-shot segmentation on PASCAL-$5^i$ dataset in Figure~\ref{fig:vis}. As detailed in Section~\ref{sec:notation}, our model is realized in the weakly supervised setting and does not observe ground truth masks of support images during training (i.e., column 1 in Figure~\ref{fig:vis}). Instead, given each image and its image-level label (i.e., class name), we first generate its pseudo mask, which serves as the guidance for our meta-learning framework. As evident in column 2 of Figure~\ref{fig:vis}, our generated pseudo masks do not cover the entire foreground object, but instead contain only discriminative regions that are informative in classifying image-level labels (e.g., muzzle of a horse, or wheels and pedals of a motorcycle). Nevertheless, from this figure, we see that our proposed model is able to predict masks for query images with satisfactory performances (e.g., columns 4 and 6 in Figure~\ref{fig:vis}). It is also worth noting that no post processing is performed on our predicted results. 

From the split-wise mean-IoUs listed in Table~\ref{tab:pascal_coco}, we see that some data splits would generally suffer from performance drops across different models including ours. In Figure~\ref{fig:fail}, we show failure segmentation example results by our weakly supervised model and the fully supervised PFENet~\cite{pfenet}. As evident in this figure, the ground truth images of such categories generally contain small foreground regions (e.g., bottle), or those that cannot be easily distinguished from the background (e.g., chair). For such image categories, training with few-shot samples would \textit{not} be expected to address semantic segmentation tasks well. This would be the limitation of developing solutions for few-shot semantic segmentation.

%% file: tabs/ablation.tex
\begin{table}[]
  \centering
  \resizebox{.9\linewidth}{!}{%
  \begin{tabular}{ccccccc} 
    \hline\hline
    \multirow{2}{*}{Method} & \multicolumn{2}{c}{\textbf{1-shot}} & \multicolumn{2}{c}{\textbf{5-shot}}  \\ 
    \cmidrule(lr){2-3}\cmidrule(lr){4-5}
    {} & \textbf{1-way} & \textbf{2-way} & \textbf{1-way} & \textbf{2-way} \\ 
    \Xhline{2\arrayrulewidth}
    \phantom{xxx}Ours w/o $E$\phantom{xxx} & 30.0 & 19.7 & 31.5 & 25.4 \\
    Ours w/o Saliency & 41.0 & 35.4 & 42.1 & 39.4 \\
    Full Version & 42.4 & 37.7 & 45.5 & 43.0 \\
    \hline\hline
    \\
  \end{tabular}
  }
  \caption{Ablation study of our model on PASCAL-$5^i$ in mean-IoUs. Ours w/o $E$ denotes our framework without meta-learner encoder~$E$, while w/o Saliency indicates our model without applying saliency gating to process the generated pseudo masks.}
  \label{tab:ablation}
\end{table}

%% file: 5_conclusion.tex
\section{Conclusion}

In this paper, we proposed a unique learning framework for few-shot semantic segmentation in a weakly supervised manner, which only observes image-level labels during both training and testing. Under such weak supervision, our proposed model serves as a pixel-level meta-learner, which produces pseudo segmentation masks for guiding pixel-wise classification in a meta-learning fashion. Our experiments confirmed the effectiveness of our model, which achieved satisfactory results on two benchmark datasets under fully supervised settings, while surpassing the state-of-the-art methods in weakly supervised settings. Finally, we point out the challenge and limitation of the task of few-shot semantic segmentation.

\paragraph{Acknowledgement} This work is supported in part by the
Ministry of Science and Technology of Taiwan under grants MOST 110-2634-F-002-036 and 110-2221-E-002-121.